\documentclass[11pt,letterpaper]{article}
\usepackage{naaclhlt2016}
\usepackage{times}
\usepackage{latexsym}
\usepackage[english]{babel}

\usepackage{graphicx} 
\graphicspath{{figures/}} 

\usepackage{amsmath} 

\usepackage{booktabs}
\usepackage{multirow}

\naaclfinalcopy 
\def\naaclpaperid{***} 

\title{Semi-supervised Clustering for Short Text via Deep Representation Learning}

 \author{Zhiguo Wang \and Haitao Mi \and Abraham Ittycheriah \\
          IBM T.J. Watson Research Center \\ Yorktown Heights, NY, USA \\ {\tt \{zhigwang, hmi, abei\}@us.ibm.com}}
\date{}

\begin{document}

\maketitle

\begin{abstract}
In this work, we propose a semi-supervised method for short text clustering,
where we represent texts as distributed vectors with neural networks, 
and use a small amount of labeled data to specify our intention for clustering.
We design a novel objective to combine the representation learning process and the k-means clustering process together, 
and optimize the objective with both labeled data and unlabeled data iteratively until convergence through three steps: 
(1) assign each short text to its nearest centroid based on its representation from the current neural networks; 
(2) re-estimate the cluster centroids based on cluster assignments from step (1); 
(3) update neural networks according to the objective by keeping centroids and cluster assignments fixed. 
Experimental results on four datasets show that our method works significantly better than several other text clustering methods.

\end{abstract}

\section{Introduction}
Text clustering is a fundamental problem in text mining and information retrieval. 
Its task is to group similar texts together such that texts within a cluster 
are more similar to texts in other clusters. 
Usually, a text is represented as a bag-of-words or 
term frequency-inverse document frequency (TF-IDF) vector, 
and then the k-means algorithm~\cite{macqueen1967some} is performed 
to partition a set of texts into homogeneous groups. 

However, when dealing with short texts, the characteristics of short text and clustering task 
raise several issues for the conventional unsupervised clustering algorithms. 
First, the number of uniqe words in each short text is small, 
as a result, the lexcical sparsity issue usually 
leads to poor clustering quality~\cite{dhillon+:2003}.
Second, for a specific short text clustering task, 
we have prior knowledge or paticular intenstions before clustering,
while fully unsupervised approaches may learn some classes the other way around.
Take the sentences in Table~\ref{tab:example} for example, 
those sentences can be clustered into different partitions based on different intentions:
\emph{apple} \{a, b, c\} and \emph{orange} \{d, e, f\} with a fruit type intension, 
or \emph{what-question} \{a, d\}, \emph{when-question} \{b, e\}, 
and \emph{yes/no-question} cluster \{c, f\} with a question type intension.

\begin{table}[tbp]
\centering
\begin{tabular}[c]{|l|}
\hline
(a) What's the color of apples? \\
(b) When will this apple be ripe? \\
(c) Do you like apples? \\
(d) What's the color of oranges? \\
(e) When will this orange be ripe? \\
(f) Do you like oranges? \\
\hline
\end{tabular}
\caption{Examples for short text clustering.}
\label{tab:example}
\end{table}

To address the lexical sparity issue, 
one direction is to enrich text representations by extracting features and relations 
from Wikipedia~\cite{banerjee2007clustering} or an ontology~\cite{fodeh2011ontology}.
But this approach requires the annotated knowlege, which is also language dependent. 
So the other direction, which directly encode texts into distributed vectors with
neural networks~\cite{hinton2006reducing,xu2015short}, becomes more interesing. 
To tackle the second problem, 
semi-supervised approaches (e.g. \cite{bilenko2004integrating,davidson2007survey,bair2013semi})
have gained significant popularity in the past decades.
Our question is can we have a unified model to integrate netural networks into the semi-supervied framework?

In this paper, we propose a unified framework for the short text clustering task.
We employ a deep neural network model to represent short sentences,
and integrate it into a semi-supervised algorithm. 
Concretely, 
we extend the objective in the classical unsupervised k-means algorithm by adding a penalty term from labeled data. 
Thus, the new objective covers three key groups of parameters: 
centroids of clusters, the cluster assignment for each text, and the parameters within deep neural networks. 
In the training procedure, we start from random initialization of centroids and neural networks, 
and then optimize the objective iteratively through three steps until converge: 
\begin{enumerate}
\item[(1)] assign each short text to its nearest centroid based on its representation from the current neural networks; 
\item[(2)] re-estimate cluster centroids based on cluster assignments from step (1); 
\item[(3)] update neural networks according to the objective by keeping centroids and cluster assignments fixed. 
\end{enumerate}
Experimental results on four different datasets show that 
our method achieves significant improvements over several other text clustering methods.

In following parts, we first describe our neural network models for text representaion (Section~\ref{sec:representation_learning}). 
Then we introduce our semi-supervised clustering method and the learning algorithm (Section~\ref{sec:semi-algo}).
Finally, we evaluate our method on four different datasets (Section~\ref{sec:exps}).


\section{Representation Learning for Short Texts}
\label{sec:representation_learning}
We represent each word with a dense vector $w$,
so that a short text $s$ is first represented as a matrix $S = [w_1,...,w_{|s|}]$, 
which is a concatenation of all vectors of $w$ in $s$, $|s|$ is the length of $s$.
Then we design two different types of neural networks to ingest the word vector sequence $S$: 
the convolutional neural networks (CNN) and the long short-term memory (LSTM).
More formally, we define the presentation function as $x = f(s)$, where $x$ is 
the represent vector of the text $s$. 
We test two encoding functions (CNN and LSTM) in our experiments.

\begin{figure}[tbp]
\begin{center}
\includegraphics[width=0.4\textwidth]{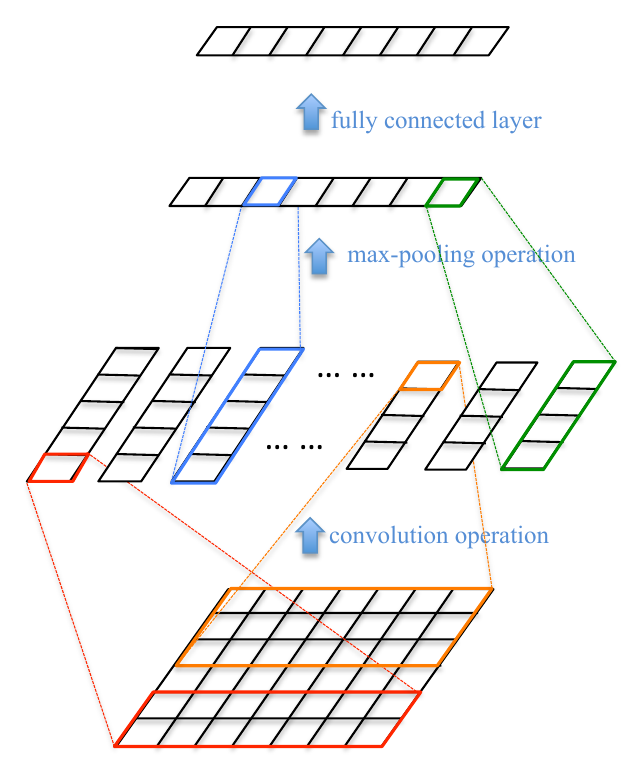}
\end{center}
\caption{CNN for text representation learning.}
\label{fig:cnn_graph}
\end{figure}

Inspired from \newcite{kim2014convolutional}, our CNN model views the sequence of word vectors as a matrix, 
and applies two sequential operations: \emph{convolution} and \emph{max-pooling}. 
Then, a fully connected layer is employed to convert the final representation vector into a fixed size. 
Figure~\ref{fig:cnn_graph} gives the diagram of the CNN model. 
In the \emph{convolution} operation, we define a list of filters \{\textbf{$w_o$}\}, 
where the shape of each filter is $d\times h$, $d$ is the dimension of word vectors and $h$ is the window size. 
Each filter is applied to a patch (a window size $h$ of vectors) of $S$, and generates a feature. 
We apply this filter to all possible patches in $S$, and produce a series of features. 
The number of features depends on the shape of the filter $w_o$ and the length of the input short text. 
To deal with variable feature size, 
we perform a \emph{max-pooling} operation over all the features to select the maximum value. 
Therefore, after the two operations, each filter generates only one feature. 
We define several filters by varying the window size and the initial values. 
Thus, a vector of features is captured after the \emph{max-pooling} operation, 
and the feature dimension is equal to the number of filters.

\begin{figure}[tbp]
\begin{center}
\includegraphics[width=0.4\textwidth]{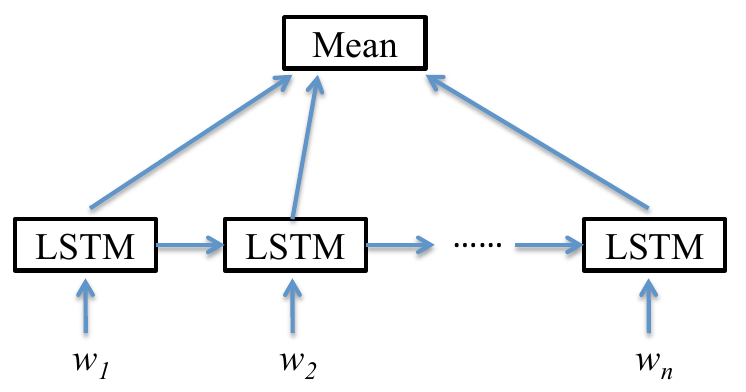}
\end{center}
\caption{LSTM for text representation learning.}
\label{fig:lstm_graph}
\end{figure}

Figure \ref{fig:lstm_graph} gives the diagram of our LSTM model. 
We implement the standard LSTM block described in \newcite{graves2012supervised}. 
Each word vector is fed into the LSTM model sequentially, 
and the mean of the hidden states over the entire sentence is taken as the final representation vector.

\section{Semi-supervised Clustering for Short Texts}
\label{sec:semi-algo}
\subsection{Revisiting K-means Clustering}
Given a set of texts $\{s_1,s_2,...,s_N\}$, 
we represent them as a set of data points $\{x_1,x_2,...,x_N\}$, 
where $x_i$ can be a bag-of-words or TF-IDF vector in traditional approaches, or a dense vector in Section~\ref{sec:representation_learning}. 
The task of text clustering is to partition the data set into some number $K$ of clusters, such that the sum of the squared distance of each data point to its closest cluster centroid is minimized. For each data point $x_n$, we define a set of binary variables $r_{nk}\in\{0,1\}$, where $k \in \{1,...,K\}$ describing which of the K clusters $x_n$ is assigned to. So that if $x_n$ is assigned to cluster $k$, then $r_{nk}=1$, and $r_{nj}=0$ for $j\neq k$. Let's define $\mu_k$ as the centroid of the $k$-th cluster. We can then formulate the objective function as
\begin{equation}
J_{unsup}=\sum_{n=1}^{N}\sum_{k=1}^{K} r_{nk}\|x_n-\mu_k\|^2
\label{equ:unsup_obj}
\end{equation}
Our goal is the find the values of $\{r_{nk}\}$ and $\{\mu_k\}$ so as to minimize $J_{unsup}$.

The k-means algorithm optimizes $J_{unsup}$ through the gradient descent approach, and results in an iterative procedure \cite{bishop2006pattern}. Each iteration involves two steps: \emph{E-step} and \emph{M-step}. In the \emph{E-step}, the algorithm minimizes $J_{unsup}$ with respect to $\{r_{nk}\}$ by keeping $\{\mu_k\}$ fixed. $J_{unsup}$ is a linear function for $\{r_{nk}\}$, so we can optimize for each data point separately by simply assigning the $n$-th data point to the closest cluster centroid. In the \emph{M-step}, the algorithm minimizes $J_{unsup}$ with respect to $\{\mu_k\}$ by keeping $\{r_{nk}\}$ fixed. $J_{unsup}$ is a quadratic function of $\{\mu_k\}$, and it can be minimized by setting its derivative with respect to $\{\mu_k\}$ to zero. 
\begin{equation}
\frac{\partial J_{unsup}}{\partial \mu_k}=2\sum_{n=1}^{N}r_{nk}(x_n-\mu_k)=0
\end{equation}
Then, we can easily solve $\{\mu_k\}$ as
\begin{equation}
\mu_k=\frac{\sum_{n=1}^N r_{nk}x_n}{\sum_{n=1}^N r_{nk}}
\end{equation}
In other words, $\mu_k$ is equal to the mean of all the data points assigned to cluster $k$.

\subsection{Semi-supervised K-means with Neural Networks}

The classical k-means algorithm only uses unlabeled data, and solves the clustering problem under the unsupervised learning framework. 
As already mentioned, the clustering results may not be consistent to our intention. 
In order to acquire useful clustering results, some supervised information should be introduced into the learning procedure. 
To this end, we employ a small amount of labeled data to guide the clustering process. 


\begin{table*}[tbp]
\centering
\begin{tabular}[c]{|l|}
\hline
1. Initialize $\{\mu_k\}$ and $f(\cdot)$. \\
2. \emph{assign\_cluster}: Assign each text to its nearest cluster centroid. \\
3. \emph{estimate\_centroid}: Estimate the cluster centroids based on the cluster assignments from step 2. \\
4. \emph{update\_parameter}: Update parameters in neural networks.\\
5. Repeat step 2 to 4 until convergence. \\
\hline
\end{tabular}
\caption{Pseudocode for semi-supervised clustering}
\label{tab:pseudocode}
\end{table*}

Following Section~\ref{sec:representation_learning}, 
we represent each text $s$ as a dense vector $x$ via neural networks $f(s)$.
Instead of training the text representation model separately, 
we integrate the training process into the k-means algorithm, 
so that both the labeled data and the unlabeled data can be used for representation learning and text clustering. 
Let us denote the labeled data set as \{$(s_1, y_1), (s_2, y_2), ..., (s_L, y_L)$\}, 
and the unlabeled data set as \{$s_{L+1}, s_{L+2},..., s_N$\}, 
where $y_i$ is the given label for $s_i$. We then define the objective function as:

\begin{equation}
\begin{split}
&J_{semi}=\alpha\sum_{n=1}^{N}\sum_{k=1}^{K} r_{nk}\|f(s_n)-\mu_k\|^2 \\
&+(1-\alpha)\sum_{n=1}^{L}\{\|f(s_n)-\mu_{g_n}\|^2 +\\
&\sum_{j\neq g_n} [l+\|f(s_n)-\mu_{g_n}\|^2-\|f(s_n)-\mu_j\|^2]_+\}
\end{split}
\label{equ:semi_obj}
\end{equation}
The objective function contains two terms. The first term is adapted from the unsupervised k-means algorithm in Eq. (\ref{equ:unsup_obj}), and the second term is defined to encourage labeled data being clustered in correlation with the given labels. $\alpha \in [0, 1]$ is used to tune the importance of unlabeled data. The second term contains two parts. The first part penalizes large distance between each labeled instance and its correct cluster centroid, where $g_n=G(y_n)$ is the cluster ID mapped from the given label $y_n$, and the mapping function $G(\cdot)$ is implemented with the Hungarian algorithm \cite{munkres1957algorithms}. The second part is denoted as a hinge loss with a margin $l$, where $[x]_+=max(x, 0)$. This part incurs some loss if the distance to the correct centroid is not shorter (by the margin $l$) than distances to any of incorrect cluster centroids.

There are three groups of parameters in $J_{semi}$: the cluster assignment of each text \{$r_{nk}$\}, the cluster centroids \{$\mu_k$\}, and the parameters within the neural network model $f(\cdot)$. Our goal is the find the values of $\{r_{nk}\}$, $\{\mu_k\}$ and parameters in $f(\cdot)$, so as to minimize $J_{semi}$. Inspired from the k-means algorithm, we design an algorithm to successively minimize $J_{semi}$ with respect to $\{r_{nk}\}$, \{$\mu_k$\}, and parameters in $f(\cdot)$. Table \ref{tab:pseudocode} gives the corresponding pseudocode. First, we initialize the cluster centroids $\{\mu_k\}$ with the \texttt{k-means++} strategy \cite{arthur2007k}, and randomly initialize all the parameters in the neural network model. Then, the algorithm iteratively goes through three steps (\emph{assign\_cluster}, \emph{estimate\_centroid}, and \emph{update\_parameter}) until $J_{semi}$ converges. 

The \emph{assign\_cluster} step minimizes $J_{semi}$ with respect to \{$r_{nk}$\} by keeping $f(\cdot)$ and $\{\mu_k\}$ fixed. Its goal is to assign a cluster ID for each data point. We can see that the second term in Eq. (\ref{equ:semi_obj}) has no relation with $\{r_{nk}\}$. Thus, we only need to minimize the first term by assigning each text to its nearest cluster centroid, which is identical to the \emph{E-step} in the k-means algorithm. In this step, we also calculate the mappings between the given labels \{$y_i$\} and the cluster IDs (with the Hungarian algorithm) based on cluster assignments of all labeled data.

The \emph{estimate\_centroid} step minimizes $J_{semi}$ with respect to $\{\mu_k\}$ by keeping $\{r_{nk}\}$ and $f(\cdot)$ fixed, which corresponds to the \emph{M-step} in the k-means algorithm. It aims to estimate the cluster centroids \{$\mu_k$\} based on the cluster assignments $\{r_{nk}\}$ from the \emph{assign\_cluster} step. The second term in Eq. (\ref{equ:semi_obj}) makes each labeled instance involved in the estimating process of cluster centroids. By solving $\partial J_{semi} / \partial \mu_k=0$, we get  

\begin{equation}
\mu_k=\frac{\sum_{n=1}^N \alpha r_{nk}f(s_n) + \sum_{n=1}^Lw_{nk}f(s_n)}{\sum_{n=1}^N \alpha r_{nk} +\sum_{n=1}^Lw_{nk}}\\
\label{equ:weight}
\end{equation}

\begin{equation}
\begin{split}
&w_{nk}=(1-\alpha)(I_{nk}^{'} + \sum_{j\neq g_n}I_{nkj}^{''} - \sum_{j\neq g_n}I_{nkj}^{'''}) \\
&I_{nk}^{'}=\delta(k,g_n)\\
&I_{nkj}^{''}=\delta(k,j)\cdot \delta_{nj}^{'}\\
&I_{nkj}^{'''}=(1-\delta(k,j))\cdot\delta_{nj}^{'}\\
&\delta_{nj}^{'}=\delta(l+\|f(s_n)-\mu_{g_n}\|^2-\|f(s_n)-\mu_j\|^2>0)
\end{split}
\label{equ:weight2}
\end{equation}
where $\delta(x_1,x_2)$=1 if $x_1$ is equal to $x_2$, otherwise $\delta(x_1,x_2)$=0, and $\delta(x)$=1 if $x$ is true, otherwise $\delta(x)$=0. The first term in the numerator of Eq. (\ref{equ:weight}) is the contributions from all data points, and $\alpha r_{nk}$ is the weight of $s_n$ for $\mu_k$. The second term is acquired from labeled data, and $w_{nk}$ is the weight of a labeled instance $s_n$ for $\mu_k$. 

The \emph{update\_parameter} step minimizes $J_{semi}$ with respect to $f(\cdot)$ by keeping $\{r_{nk}\}$ and $\{\mu_k\}$ fixed, which has no counterpart in the k-means algorithm. The main goal is to update parameters for the text representation model. We take $J_{semi}$ as the loss function, and train neural networks with the \emph{Adam} algorithm \cite{kingma2014adam}.

\section{Experiment}
\label{sec:exps}
\subsection{Experimental Setting}
We evaluate our method on four short text datasets. (1) \emph{question\_type} is the TREC question dataset \cite{li2002learning}, where all the questions are classified into 6 categories: abbreviation, description, entity, human, location and numeric. (2) \emph{ag\_news} dataset contains short texts extracted from the AG's news corpus, where all the texts are classified into 4 categories: World, Sports, Business, and Sci/Tech \cite{zhang2015text}. (3) \emph{dbpedia} is the DBpedia ontology dataset, which is constructed by picking 14 non-overlapping classes from DBpedia 2014 \cite{lehmann2014dbpedia}. (4) \emph{yahoo\_answer} is the 10 topics classification dataset extracted from Yahoo! Answers Comprehensive Questions and Answers version 1.0 dataset by \newcite{zhang2015text}. We use all the 5,952 questions for the \emph{question\_type} dataset. But the other three datasets contain too many instances (e.g. 1,400,000 instances in yahoo\_answer). Running clustering experiments on such a large dataset is quite inefficient. Following the same solution in \cite{xu2015short}, 
we randomly choose 1,000 samples for each classes individually for the other three datasets. Within each dataset, we randomly sample 10\% of the instances as labeled data, and evaluate the performance on the remaining 90\% instances. Table \ref{tab:statistics} summarizes the statistics of these datasets.

\begin{table}[tbp]
\centering
\begin{tabular}{lrrr}
\toprule
dataset        & class\# & total\# & labeled\# \\
\midrule
question\_type & 6       & 5,953   & 595       \\
ag\_news       & 4       & 4,000   & 400       \\
dbpedia        & 14      & 14,000  & 1,400     \\
yahoo\_answer  & 10      & 10,000  & 1,000    \\
\bottomrule
\end{tabular}
\caption{Statistics for the short text datasets}
\label{tab:statistics}
\end{table}

In all experiments, we set the size of word vector dimension as $d$=300 \footnote{We tuned different dimensions for word vectors. When the size is small (50 or 100), performance drops significantly. When the size is larger (300, 500 or 1000), the curve flattens out. To make our model more efficient, we fixed it as 300.}, and pre-train the word vectors with the \emph{word2vec} toolkit \cite{mikolov2013efficient} on the English Gigaword (LDC2011T07). The number of clusters is set to be the same number of labels in the dataset. The clustering performance is evaluated with two metrics: Adjusted Mutual Information (AMI) \cite{vinh2009information} and accuracy (ACC) \cite{amigo2009comparison}. In order to show the statistical significance, the performance of each experiment is the average of 10 trials.

\subsection{Model Properties}
There are several hyper-parameters in our model, e.g., the output dimension of the text representation models, and the $\alpha$ in Eq. (\ref{equ:semi_obj}). The choice of these hyper-parameters may affect the final performance. In this subsection, we present some experiments to demonstrate the properties of our model, and find a good configuration that we use to evaluate our final model. All the experiments in this subsection were performed on the \emph{question\_type} dataset.

\begin{figure}[tbp]
\begin{center}
\includegraphics[width=0.45\textwidth]{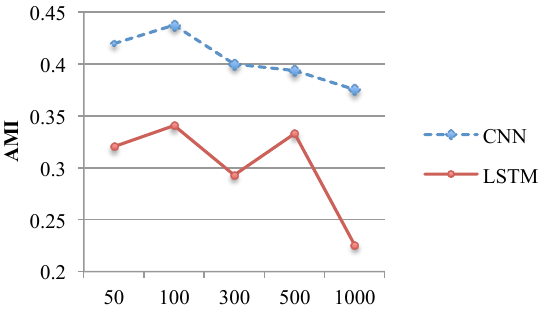}
\end{center}
\caption{Influence of the short text representation model, where the x-axis is the output dimension of the text representation models.}
\label{fig:num_dims}
\end{figure}

First, we evaluated the effectiveness of the output dimension in text representation models. We switched the dimension size among \{50, 100, 300, 500, 1000\}, and fixed the other options as: $\alpha=0.5$, the filter types in the CNN model including \{unigram, bigram, trigram\} and 500 filters for each type. Figure \ref{fig:num_dims} presents the AMIs from both CNN and LSTM models. We found that 100 is the best output dimension for both CNN and LSTM models. Therefore, we set the output dimension as 100 in the following experiments.

\begin{figure}[tbp]
\begin{center}
\includegraphics[width=0.45\textwidth]{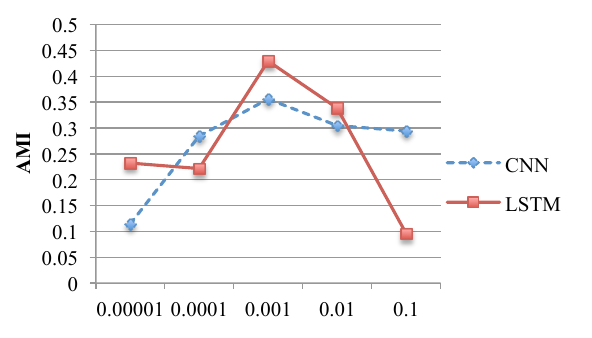}
\end{center}
\caption{Influence of unlabeled data, where the x-axis is $\alpha$ in Eq. (\ref{equ:semi_obj}).}
\label{fig:alphas}
\end{figure}

Second, we studied the effect of $\alpha$ in Eq. (\ref{equ:semi_obj}), which tunes the importance of unlabeled data. We varied $\alpha$ among \{0.00001, 0.0001, 0.001, 0.01, 0.1\}, and remain the other options as the last experiment. Figure \ref{fig:alphas} shows the AMIs from both CNN and LSTM models. We found that the clustering performance is not good when using a very small $\alpha$. By increasing the value of $\alpha$, we acquired progressive improvements, and reached to the peak point at $\alpha$=0.01. After that, the performance dropped. Therefore, we choose $\alpha$=0.01 in the following experiments. This results also indicate that the unlabeled data are useful for the text representation learning process. 

\begin{figure}[tbp]
\begin{center}
\includegraphics[width=0.45\textwidth]{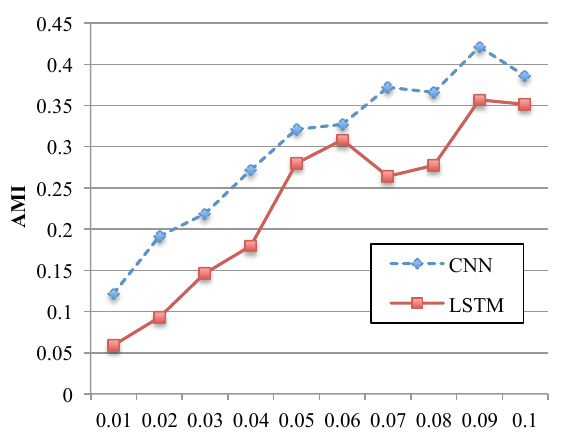}
\end{center}
\caption{Influence of labeled data, where the x-axis is the ratio of data with given labels.}
\label{fig:ratios}
\end{figure}

Third, we tested the influence of the size of labeled data. We tuned the ratio of labeled instances from the whole dataset among [1\%, 10\%], and kept the other configurations as the previous experiment. The AMIs are shown in Figure \ref{fig:ratios}. We can see that the more labeled data we use, the better performance we get. Therefore, the labeled data are quite useful for the clustering process.

\begin{figure}[tbp]
\begin{center}
\includegraphics[width=0.45\textwidth]{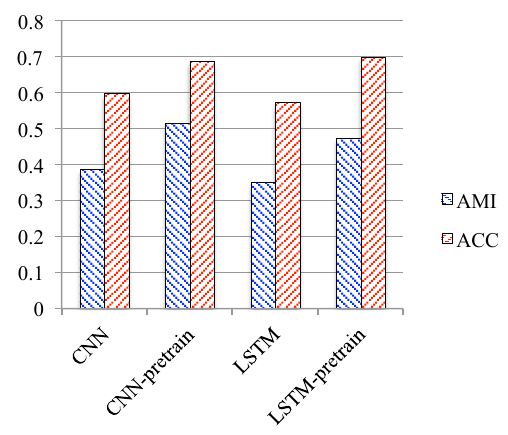}
\end{center}
\caption{Influence of the pre-training strategy.}
\label{fig:pretrain}
\end{figure}

Fourth, we checked the effect of the pre-training strategy for our models. We added a softmax layer on top of our CNN and LSTM models, where the size of the output layer is equal to the number of labels in the dataset. We then trained the model through the classification task using all labeled data. After this process, we removed the top layer, and used the remaining parameters to initialize our CNN and LSTM models. The performance for our models with and without pre-training strategy are given in Figure \ref{fig:pretrain}. We can see that the pre-training strategy is quite effective for our models. Therefore, we use the pre-training strategy in the following experiments.

\subsection{Comparing with other Models}

\begin{table*}[tbp]
\centering
\begin{tabular}{cl|cc|cc|cc|cc}
\toprule
\multicolumn{2}{l}{} & \multicolumn{2}{|c}{question\_type} & \multicolumn{2}{|c}{ag\_news}   & \multicolumn{2}{|c}{dbpedia}    & \multicolumn{2}{|c}{yahoo\_answer} \\ 
\cline{3-10}  
\multicolumn{2}{l|}{}  & AMI & ACC & AMI & ACC & AMI & ACC & AMI & ACC \\
\midrule
\multirow{3}{*}{Unsup.} 
& bow & 0.028 & 0.257 & 0.029 & 0.311 & 0.578 & 0.546 & 0.019 & 0.140 \\
& tf-idf & 0.031 & 0.259 & 0.168 & 0.449 & 0.558 & 0.527 & 0.023 & 0.145 \\ 
& average-vec & 0.135 & 0.356 & 0.457 & 0.737 & 0.610 & 0.619 & 0.077 & 0.222 \\ 
\midrule
\multirow{7}{*}{Sup.} 
& metric-learn-bow & 0.104 & 0.380 & 0.459 & 0.776 & 0.808 & 0.854 & 0.125 & 0.329 \\ 
& metric-learn-idf & 0.114 & 0.379 & 0.443 & 0.765 & 0.821 & 0.876 & 0.150 & 0.368 \\ 
& metric-learn-ave-vec & 0.304 & 0.553 & 0.606 & 0.851 & 0.829 & 0.879 & 0.221 & 0.400 \\ 
& cnn-classifier & 0.511 & \textbf{0.771} & 0.554 & 0.771 & 0.879 & 0.938 & 0.285 & 0.501 \\ 
& cnn-represent.  & 0.442 & 0.618 & 0.604 & 0.833 & 0.864 & 0.899 & 0.210 & 0.334 \\ 
& lstm-classifier & 0.482 & 0.741 & 0.524 & 0.763 & 0.862 & 0.928 & 0.283 & 0.512 \\ 
& lstm-represent. & 0.421 & 0.618 & 0.535 & 0.771 & 0.667 & 0.706 & 0.152 & 0.272 \\ 
\midrule
\multirow{2}{*}{Semisup.} 
& semi-cnn & \textbf{0.529} & 0.739 & \textbf{0.662} & \textbf{0.876} & \textbf{0.894} & \textbf{0.945} & \textbf{0.338} & \textbf{0.554} \\
& semi-lstm & 0.492 & 0.712 & 0.599 & 0.830 & 0.788 & 0.802 & 0.187 & 0.337 \\ 
\bottomrule
\end{tabular}
\caption{Performance of all systems on each dataset.}
\label{tab:compare}
\end{table*}

\begin{figure*}[tbp]
\begin{center}
\includegraphics[width=1.0\textwidth]{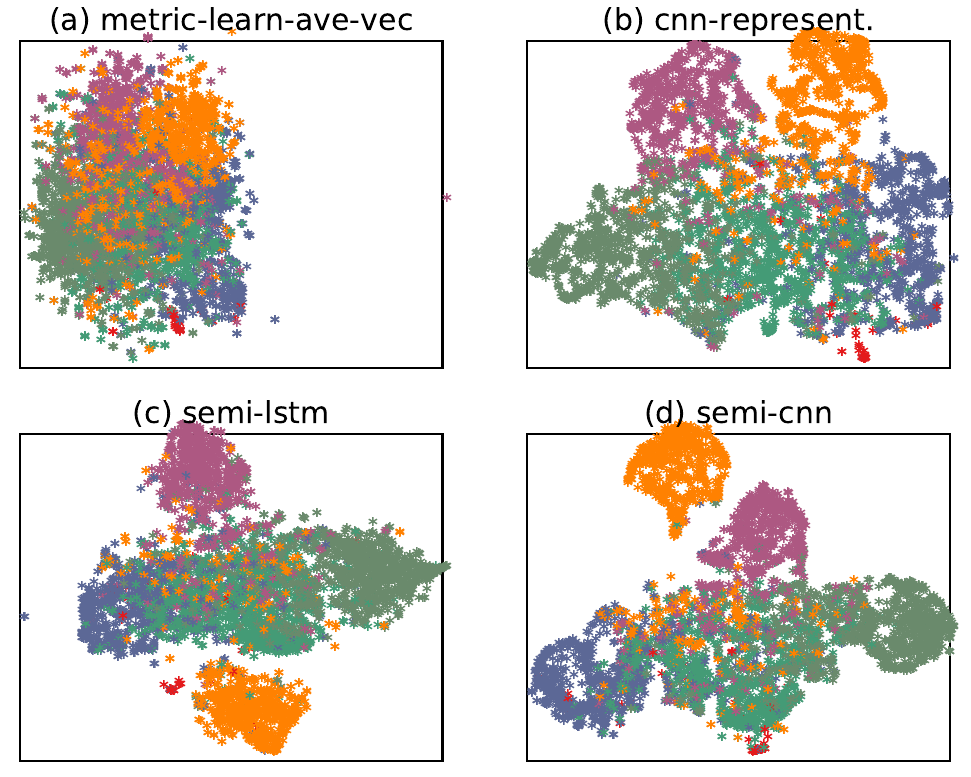}
\end{center}
\caption{t-SNE visualizations of clustering results.}
\label{fig:all_clusters}
\end{figure*}

In this subsection, we compared our method with some representative systems. We implemented a series of clustering systems. All of these systems are based on the k-means algorithm, but they represent short texts differently:
\begin{description}
  \item [bow] represents each text as a bag-of-words vector.
  \item [tf-idf] represents each text as a TF-IDF vector.
  \item [average-vec] represents each text with the average of all word vectors within the text.
  \item [metric-learn-bow] employs the metric learning method proposed by \newcite{weinberger2005distance}, and learns to project a bag-of-words vector into a 300-dimensional vector based on labeled data. 
  \item [metric-learn-idf] uses the same metric learning method, and learns to map a TF-IDF vector into a 300-dimensional vector based on labeled data.
  \item [metric-learn-ave-vec] also uses the metric learning method, and learns to project an averaged word vector into a 100-dimensional vector based on labeled data. 
\end{description}

We designed two classifiers (cnn-classifier and lstm-classifier) by adding a softmax layer on top of our CNN and LSTM models. We trained these two classifiers with labeled data, and utilized them to predict labels for unlabeled data. We also built two text representation models (``cnn-represent." and ``lstm-represent.") by
setting parameters of our CNN and LSTM models with the corresponding parameters in cnn-classifier and lstm-classifier. Then, we used them to represent short texts into vectors, and applied the k-means algorithm for clustering.

Table \ref{tab:compare} summarizes the results of all systems on each dataset, where ``semi-cnn" is our semi-supervised clustering algorithm with the CNN model, and ``semi-lstm" is our semi-supervised clustering algorithm with the LSTM model. We grouped all the systems into three categories: unsupervised (Unsup.), supervised (Sup.), and semi-supervised (Semisup.) \footnote{All clustering systems are based on the same number of instances (total\# in Table \ref{tab:statistics}). For the semi-supervised and supervised systems, the labels for 1\% of the instances are given (labeled\# in Table \ref{tab:statistics}). And the evaluation was conducted only on the unlabeled portion.}. We found that the supervised systems worked much better than the unsupervised counterparts, which implies that the small amount of labeled data is necessary for better performance. We also noticed that within the supervised systems, the systems using deep learning (CNN or LSTM) models worked  better than the systems using metric learning method, which shows the power of deep learning models for short text modeling. Our ``semi-cnn" system got the best performance on almost all the datasets. 

Figure \ref{fig:all_clusters} visualizes clustering results on the \emph{question\_type} dataset from four representative systems. In Figure \ref{fig:all_clusters}(a), clusters severely overlap with each other.
When using the CNN sentence representation model, we can clearly identify all clusters in Figure \ref{fig:all_clusters}(b), but the boundaries between clusters are still obscure. The clustering results from our semi-supervised clustering algorithm are given in Figure \ref{fig:all_clusters}(c) and Figure \ref{fig:all_clusters}(d). We can see that the boundaries between clusters become much clearer. Therefore, our algorithm is very effective for short text clustering.

\section{Related Work}
Existing semi-supervised clustering methods fall into two categories: \emph{constraint-based} and \emph{representation-based}. In \emph{constraint-based} methods \cite{davidson2007survey}, some labeled information is used to constrain the clustering process. In \emph{representation-based} methods \cite{bair2013semi}, a representation model is first trained to satisfy the labeled information, and all data points are clustered based on representations from the representation model. \newcite{bilenko2004integrating} proposed to integrate there two methods into a unified framework, which shares the same idea of our proposed method. However, they only employed the metric learning model for representation learning, which is a linear projection. Whereas, our method utilized deep learning models to learn representations in a more flexible non-linear space. \newcite{xu2015short} also employed deep learning models for short text clustering. However, their method separated the representation learning process from the clustering process, so it belongs to the \emph{representation-based} method. Whereas, our method combined the representation learning process and the clustering process together, and utilized both labeled data and unlabeled data for representation learning and clustering.  

\section{Conclusion}
In this paper, we proposed a semi-supervised clustering algorithm for short texts. We utilized deep learning models to learn representations for short texts, and employed a small amount of labeled data to specify our intention for clustering. We integrated the representation learning process and the clustering process into a unified framework, so that both of the two processes get some benefits from labeled data and unlabeled data. Experimental results on four datasets show that our method is more effective than other competitors.


\bibliographystyle{naaclhlt2016}
\bibliography{cluster}

\end{document}